\documentclass[letterpaper, 10 pt, journal, twoside]{IEEEtran}
\let\authorrefmark\IEEEauthorrefmark

\usepackage{style}
\begin{document}
\maketitle

\begin{abstract}
This paper presents a theoretical overview of a Neural Contraction Metric (NCM): a neural network model of an optimal contraction metric and corresponding differential Lyapunov function, the existence of which is a necessary and sufficient condition for incremental exponential stability of non-autonomous nonlinear system trajectories. Its innovation lies in providing formal robustness guarantees for learning-based control frameworks, utilizing contraction theory as an analytical tool to study the nonlinear stability of learned systems via convex optimization. In particular, we rigorously show in this paper that, by regarding modeling errors of the learning schemes as external disturbances, the NCM control is capable of obtaining an explicit bound on the distance between a time-varying target trajectory and perturbed solution trajectories, which exponentially decreases with time even under the presence of deterministic and stochastic perturbation. These useful features permit simultaneous synthesis of a contraction metric and associated control law by a neural network, thereby enabling real-time computable and probably robust learning-based control for general control-affine nonlinear systems.
\end{abstract}
\section{Introduction}
\label{introduction}
Since the development of reinforcement learning and neural networks~\cite{sutton,ndp}, leveraging Artificial Intelligence (AI) and Machine Learning (ML) methods in the field of systems and control theory has been an active field of research. Although successful, conventional black-box AI and ML approaches often lack formal robustness and stability guarantees, which are essential in implementing safety-critical control schemes for, \eg{}, robotic and aerospace systems.
\subsubsection*{Related Work}
Contraction theory is one of the most powerful mathematical tools for providing such guarantees for learning-based control frameworks~\cite{contraction,tutorial}. Generalizing Lyapunov theory, it studies differential dynamics of a nonlinear system using a Riemannian contraction metric and its uniformly positive definite matrix, thereby designing a differential Lyapunov function to characterize a necessary and sufficient condition for exponential convergence of any couple of the system trajectories. Due to the differential nature of contraction theory which allows utilizing LTV system-type approaches~\cite{cvstem,ncm,nscm,ancm,tutorial}, its nonlinear stability analysis reduces to the problem of finding a contraction metric that satisfies a Linear Matrix Inequality (LMI)~\cite{lmi}.

Obtaining contraction metrics analytically is challenging for general nonlinear systems, and thus several techniques have been proposed to find them numerically using the LMI property. In particular, it is proposed in~\cite{ccm} that sufficient conditions for incremental exponential stabilizability of nonlinear systems can be expressed as convex constraints, enabling systematic synthesis of a differential feedback control. Such an idea is further extended to develop the method of ConVex optimization-based Steady-state Tracking Error Minimization (CV-STEM)~\cite{cvstem,ncm,nscm,tutorial}, which computes a contraction metric and its associated feedback control and state estimation laws via convex optimization for optimal deterministic/stochastic disturbance attenuation. 

A critical limitation of these optimization-based schemes is their computational complexity. In essence, a Neural Contraction Metric (NCM) and its extensions~\cite{ncm,chuchu,nscm,ancm,lagros,tutorial} are developed to model the CV-STEM contraction metric using a Deep Neural Network (DNN), achieving real-time implementable control, estimation, and motion planning of nonlinear systems with guaranteed robustness and stability. We again remark that such formal guarantees are difficult to quantify for black-box AI and ML techniques without considering a contracting stability property.
\subsubsection*{Contribution}
This paper presents rigorous proofs on the aforementioned theoretical guarantees of the NCM methods~\cite{ncm,chuchu,nscm,ancm,lagros,tutorial}, which provide an explicit exponential bound on the distance between a target trajectory and solution trajectories with the NCM learning-based control, even under the presence of learning errors and deterministic/stochastic perturbation. We also derive one explicit way to optimally sample NCM training data by formulating the CV-STEM, to design contracting differential feedback control that minimizes the steady-state of the computed bound for optimal disturbance attenuation.
Furthermore, we provide an algorithmic overview on DNN training for simultaneous synthesis of the contraction metric and control, to see how the NCM robustness and stability guarantees corroborate the success of learning-based control.
\subsubsection*{Notation}
\label{notation}
For $x \in \mathbb{R}^n$ and $A \in \mathbb{R}^{n \times m}$, we let $\|x\|$, $\delta x$, $\|A\|$, and $\|A\|_F$ denote the Euclidean norm, infinitesimal displacement at a fixed time, induced 2-norm, and Frobenius norm, respectively.
For $A \in \mathbb{R}^{n \times n}$, we use $A \succ 0$, $A \succeq 0$, $A \prec 0$, and $A \preceq 0$ for the positive definite, positive semi-definite, negative definite, negative semi-definite matrices, respectively, and $\sym(A) = (A+A^{\top})/2$. Also, $\mathop{\mathbb{E}}$ denotes the expected value operator.
\section{Neural Contraction Metrics for Learning-based Control~\cite{ncm,nscm,ancm,lagros,tutorial}}
\label{learning}
In this paper, we consider the following smooth non-autonomous (\ie{} time-varying) nonlinear system, perturbed by deterministic disturbances $d(x,t)$ with $\sup_{x,t}\|d(x,t)\|=\bar{d}$ and Gaussian white noise $\mathscr{W}(t)$ with $\|G(x,t)\|_F \leq \bar{g}$:
\begin{align}\label{eq:xfx}
d{x}=&{h}({x},u,t)dt+d(x,t)dt+G(x,t)d\mathscr{W}(t)
\end{align}
where $t \in \mathbb{R}_{+}$, ${x}:\mathbb{R}_{+}\mapsto\mathbb{R}^n$ is the system state, $u \in \mathbb{R}^m$ is the system control input, ${h}:
\mathbb{R}^n\times\mathbb{R}^m\times\mathbb{R}_{+}\rightarrow\mathbb{R}^n$ is a known smooth function, $d:\mathbb{R}^n\times\mathbb{R}_{+} \mapsto \mathbb{R}^{n}$ and $G:\mathbb{R}^n\times\mathbb{R}_{+} \to \mathbb{R}^{n\times w}$ are unknown bounded functions for external disturbances, $\mathscr{W}:\mathbb{R}_{+} \mapsto \mathbb{R}^{w}$ is a $w$-dimensional Wiener process, and we consider the case where $\bar{d},\bar{g} \in [0,\infty)$ are given. Let us first introduce the following definition of contraction~\cite{contraction}.
\begin{definition}
\label{def:contracting}
The system \eqref{eq:xfx} is said to be contracting if there exist a uniformly positive definite matrix ${M}({x},t)={{\Theta}}({x},t)^{T}{{\Theta}}({x},t) \succ 0,~\forall x,t$ and a feedback control law $u=u^*(x,x_d,u_d,t)$ \st{} the following condition holds $\forall x,t$:
\begin{align}
{\dot{M}}+2\sym{}\left(M\frac{\partial
{h}(x,u^*(x,x_d,u_d,t),t)}{\partial {x}}\right)+\Xi(x,t) \preceq 0
\label{eq_MdotContracting}
\end{align}
where $\exists \Xi(x,t) \succ 0,~\forall x,t$, $M=M(x,t)$, and $x_d$ is the target trajectory of \eqref{eq:xfx} given as follows:
\begin{align}
\label{target_dynamics}
\text{$\dot{x}_d={h}(x_d,u_d,t)$ \st{} $u_d=u^*(x_d,x_d,u_d,t)$}.
\end{align}
We call $M$ and $u^*$ satisfying \eqref{eq_MdotContracting} a contraction metric and contracting control law, respectively.
\end{definition}

It is shown that \eqref{eq_MdotContracting} is a necessary and sufficient condition for incremental exponential stability of the system trajectories with respect to each other~\cite{contraction,tutorial}.
\subsection{Problem Formulation}
Finding an optimal contraction metric $M$ for general nonlinear systems is challenging and could involve solving optimization problems at each time instant~\cite{cvstem,observer} (see Sec.~\ref{Sec:ncm}), which is not suitable for systems with limited computational capacity. A Neural Contraction Metric (NCM) and its extensions~\cite{ncm,nscm,ancm,lagros,tutorial} have been proposed to approximate a contracting control law $u^*$, \ie{}, a computationally-expensive control law parameterized by the contraction metric given by \eqref{eq_MdotContracting}, utilizing a Deep Neural Network (DNN).
\begin{definition}
\label{Def:NCM}
A Neural Contraction Metric (NCM) is a DNN model of a contraction metric given by Definition~\ref{def:contracting}.
\end{definition}
We consider the problem of modeling the contracting control law $u^*$ by a learning-based control law $u_L(x,x_d,u_d,t)$ parameterized by the NCM of Definition~\ref{Def:NCM} (see~\cite{cvstem,ccm} and Sec.~\ref{Sec:ncm} for how to obtain $u^*$). Let $x$ be the trajectory of \eqref{eq:xfx} controlled by $u=u_L$, and $x_d$ be the target trajectory of \eqref{target_dynamics}. Since we have ${h}(x,u_L,t) = {h}(x,u^*,t)+({h}(x,u_L,t)-{h}(x,u^*,t))$, a virtual system of $q$, which has $q=x$ and $q=x_d$ as its particular solutions, is given as follows:
\begin{align}
\label{virtual_system_general}
dq = {h}(q,u^*(q,x_d,u_d,t),t)dt+d_q(q,t)dt+G_q(q,t)d\mathscr{W}
\end{align}
where $d_q$ and $G_q$ are parameterized linearly to verify $d_q(q=x_d,t) = 0$, $d_q(q=x,t) = {h}(x,u_L(x,x_d,u_d,t),t)-{h}(x,u^*(x,x_d,u_d,t),t)+d(x,t)$, $G_q(q=x_d,t)=0$, and $G_q(q=x,t) = G(x,t)$. 
\begin{assumption}
\label{assump_learning}
In this section, we assume the following.
\setlength{\leftmargini}{5pt}     
\begin{itemize}
	\setlength{\itemsep}{1pt}      
	\setlength{\parskip}{0pt}      
    \item For all $(x,x_d,u_d,t)\in{S}$, $u_L$ is learned to satisfy
    \begin{align}
    \label{learning_cond}
    \|u_L(x,x_d,u_d,t)-u^*(x,x_d,u_d,t)\| \leq \epsilon_{\ell0}+\epsilon_{\ell1}\int^x_{x_d}\|\delta q\|
    \end{align}
    where ${S}={S}_x\times{S}_x\times{S}_u\times{S}_t$, ${S}_x\subset\mathbb{R}^n$, ${S}_u\subset\mathbb{R}^m$, and ${S}_t\subset\mathbb{R}_+$ are some given compact sets, and $\int^x_{x_d}\|\delta q\|$ is a path integral from $x_d$ to $x$ parameterized by $q$ of \eqref{virtual_system_general}.
    \item ${h}$ of \eqref{eq:xfx} is Lipschitz with respect to $u$, \ie{}, $\exists L_u \in [0,\infty)$ \st{}
    \begin{align}
    \label{Lipschitz_u}
    \|{h}(x,u,t)-{h}(x,u',t)\| \leq L_u\|u-u'\|,~\forall u,u'\in\mathbb{R}^m
    \end{align}
    for all $x\in{S}_x$ and $t\in{S}_t$.
    \item Let ${S}_{xt}={S}_x\times{S}_t$. When $G \neq 0$ in \eqref{eq:xfx}, $\partial M/\partial {x_i}$ of $M$ in \eqref{eq_MdotContracting} is Lipschitz with respect to $x$, \ie{}, $\exists L_m \in[0,\infty)$ \st{}
    \begin{align}
    \label{eq_Mlipschitz}
    \left\|\frac{\partial M}{\partial {x_i}}(x,t)-\frac{\partial M}{\partial {x_i}}(x',t)\right\|\leq L_m\|x-x'\|,~\forall (x,t),(x',t)\in{S}_{xt}.~~~
    \end{align}
\end{itemize}
\end{assumption}
\begin{remark}
\eqref{learning_cond} can be empirically guaranteed, \eg{}, by DNNs, which have been shown to generalize well to the set of unseen events that are from almost the same distribution as their training set~\cite{rethinknet,8794351}, and consequently, obtaining a tighter and more general bound for the learning error has been an active field of research~\cite{marginbounds1}. The condition \eqref{learning_cond} has thus become a common assumption in analyzing the performance of learning-based control~\cite{8794351,lagros,ancm,tutorial}.
\end{remark}
\subsection{Nonlinear Robustness and Stability Analysis}
The following theorem rigorously studies robustness and stability of the NCM-parameterized control law $u_L$.
\begin{theorem}
\label{Thm:learning1}
Suppose that Assumption~\ref{assump_learning} holds. Suppose also that the contraction metric $M(x,t)$ of \eqref{eq_MdotContracting} is bounded as
\begin{align}
\label{Mcon}
\underline{m} I \preceq M(x,t) \preceq \overline{m} I,~\forall x,t
\end{align}
for $\exists \underline{m},\overline{m} \in(0,\infty)$, and $u^*$ of \eqref{eq_MdotContracting} is given with $\Xi$ defined as
\begin{align}
\label{def:Xi}
\Xi(x,t) = 2\alpha M(x,t)+\alpha_s I
\end{align}
where $\exists\alpha\in (0,\infty)$ is the contraction rate and $\alpha_s = L_m\bar{g}^2(\alpha_G+{1}/{2})$ with an arbitrary constant $\alpha_G \in (0,\infty)$ (see \eqref{bound1}). If the learning error $\epsilon_{\ell1}$ of \eqref{learning_cond} and another arbitrary constant $\alpha_d \in (0,\infty)$ (see \eqref{bound2}) are selected to satisfy
\begin{align}
\label{epsilon_ell2_condition}
\exists \alpha_{\ell} \in(0,\infty)\text{ \st{} }\alpha_{\ell}=\alpha-(\alpha_d/2+L_u\epsilon_{\ell1}\sqrt{\overline{m}/\underline{m}}) > 0
\end{align}
then we have the following bound for all $(x,t) \in {S}_x\times{S}_t$:
\begin{align}
\label{Eq:learning_bound_general}
\mathop{\mathbb{E}}\left[\|x(t)-x_d(t)\|^2\right] \leq \frac{C}{2\alpha_{\ell}}\frac{\overline{m}}{\underline{m}}+\frac{\mathop{\mathbb{E}}[V_{s\ell}(0)]e^{-2\alpha_{\ell} t}}{\underline{m}}
\end{align}
where $V_{s\ell}(t) = \int_{x_d}^x\delta q(t)^{\top}M(q(t),t)\delta q(t)$, $C = \bar{g}^2({2}{\alpha_G}^{-1}+1)+(L_u\epsilon_{\ell0}+\bar{d})^2\alpha_d^{-1}$, $\bar{d}$, $\bar{g}$, and $q$ are given in \eqref{eq:xfx} and \eqref{virtual_system_general}, $x$ is the trajectory of \eqref{eq:xfx} controlled by $u=u_L$, and $x_d$ is the target trajectory given by \eqref{target_dynamics}.
\end{theorem}
\begin{proof}
Since $\partial M/\partial {x_i}$ is Lipschitz as in \eqref{eq_Mlipschitz}, we have $\|\partial^2 M/(\partial {x_i}\partial {x_j})\| \leq L_m$ and $\left\|\partial M/\partial {x_i}\right\| \leq \sqrt{{2L_m}{\overline{m}}}$ using \eqref{Mcon} as derived in~\cite{nscm}. Also, $d_q$ of \eqref{virtual_system_general} is bounded as
\begin{align}
\label{d_bound}
\|d_q(x,t)\| \leq L_u(\epsilon_{\ell0}+\epsilon_{\ell1}\int^x_{x_d}\|\delta q\|) + \bar{d}
\end{align}
by the Lipschitz condition \eqref{Lipschitz_u} and learning error bound \eqref{learning_cond}. Let $\mathscr{L}$ be the infinitesimal differential generator in~\cite[p. 15]{sto_stability_book}. Since we have $\int^x_{x_d}\|\delta q\|\leq V_{\ell}/\sqrt{\underline{m}}$ and $V_{\ell}^2\leq V_{s\ell}$~\cite{tutorial} for $V_{\ell}=\int_{x_d}^x\|\Theta \delta q\|$ and $M=\Theta^{\top}\Theta$, computing $\mathscr{L}V_{s\ell}$ using these bounds as in \cite{cvstem} yields
\begin{align}
&\mathscr{L}V_{s\ell} \leq \int_{x_d}^x\delta q^{\top}(\dot{M}+2\sym(M{h}_x))\delta q+2L_u\epsilon_{\ell1}\sqrt{\overline{m}/\underline{m}}V_{s\ell} \nonumber \\
\label{Vsl_LV}
&+2\sqrt{\overline{m}}d_{\epsilon}V_{\ell}+\bar{g}^2(\overline{m}+\int_{x_d}^xL_m\|\delta q\|^2/2+2\sqrt{{2L_m}{\overline{m}}}\|\delta q\|)
\end{align}
where ${h}_x=\partial {h}/\partial x$ and $d_{\epsilon} = L_u\epsilon_{\ell0} + \bar{d}$. Using the relation $2ab \leq c^{-1}a^2+c b^2$ which holds for any $a,b\in\mathbb{R}$ and $c \in (0,\infty)$, we have that 
\begin{align}
\label{bound1}
2\sqrt{\overline{m}}d_{\epsilon}\|\Theta \delta q\| \leq& \alpha_d\|\Theta \delta q\|^2+\overline{m}d_{\epsilon}^2\alpha_d^{-1} \\
\label{bound2}
2\sqrt{{2L_m}{\overline{m}}}\|\delta q\| \leq& L_m\alpha_G\|\delta q\|^2+2{\overline{m}}\alpha_G^{-1}
\end{align}
for any $\alpha_d,\alpha_G\in(0,\infty)$. Substituting \eqref{bound1} and \eqref{bound2} into \eqref{Vsl_LV} along with the contraction condition \eqref{eq_MdotContracting} for $u^*$ yields
\begin{align}
\mathscr{L}V_{s\ell} \leq& \int_{x_d}^x\delta q^{\top}(-\Xi+(\alpha_d+2L_u\epsilon_{\ell1}\sqrt{\chi})M+\alpha_sI)\delta q+{\overline{m}}{C}
\end{align}
where $\chi=\overline{m}/\underline{m}$, $\alpha_s = L_m\bar{g}^2(\alpha_G+{1}/{2})$, and $C = \bar{g}^2({2}{\alpha_G}^{-1}+1)+\bar{d}_{\epsilon}^2\alpha_d^{-1}$. Thus, by definition of $\Xi$ in \eqref{def:Xi}, we have $\mathscr{L}V_{s\ell} \leq -2\alpha_{\ell} V_{s\ell}+{\overline{m}}{C}$ due to the condition \eqref{epsilon_ell2_condition}. This results in the desired relation \eqref{Eq:learning_bound_general} due to Theorem 1 of~\cite{cvstem}.
\end{proof}

Theorem~\ref{Thm:learning1} states that, as long as we have a contracting controller $u^*$ which satisfies \eqref{eq_MdotContracting} of Definition~\ref{def:contracting}, the NCM controller $u_L$ of \eqref{learning_cond} ensures that the mean-squared distance from any target trajectory $x_d$ controlled by $u_d=u^*(x_d,x_d,u_d,t)$ to the one controlled by $u_L$ with deterministic and stochastic perturbation is exponentially bounded in time, even under the presence of the learning error $\epsilon_{\ell0}$ and $\epsilon_{\ell1}$. Such an explicit exponential bound is difficult to obtain for systems without a contraction property.

In particular, when there is no stochastic disturbance in the system \eqref{eq:xfx}, \ie{}, $G=0$, we have the following result.
\begin{theorem}
\label{Cor:learning_simple}
Suppose \eqref{learning_cond} and \eqref{Lipschitz_u} of Assumption~\ref{assump_learning} hold for $u^*$ that satisfies \eqref{eq_MdotContracting} with $\Xi = 2\alpha M(x,t)$, where $\exists \alpha \in (0,\infty)$ is the contraction rate. Now let $x_d$ be the target trajectory of \eqref{target_dynamics}, and $x$ be the trajectory controlled by $u_L$ with $G=0$. If $M$ is bounded as in \eqref{Mcon}, and if the learning error $\epsilon_{\ell1}$ of \eqref{learning_cond} is sufficiently small to satisfy
\begin{align}
\label{epsilon_ell2_condition_det}
\exists \alpha_{\ell} \in(0,\infty)\text{ \st{} }\alpha_{\ell}=\alpha-L_u\epsilon_{\ell1}\sqrt{\overline{m}/\underline{m}} > 0
\end{align}
then we have the following bound:
\begin{align}
\label{bound_no_dist}
\|x(t)-x_d(t)\| \leq \frac{\|V_{\ell}(0)\|}{\sqrt{\underline{m}}}e^{-\alpha_{\ell} t}+\frac{L_u\epsilon_{\ell0}+\bar{d}}{\alpha_{\ell}}\sqrt{\frac{\overline{m}}{\underline{m}}}(1-e^{-\alpha_{\ell} t})~~~~
\end{align}
where $V_{\ell}(t) = \int_{x_d}^{x}\|\Theta(q(t),t)\delta q(t)\|$ and $q$ is given by \eqref{virtual_system_general} with the same $d_q$ and $G_q=0$ (note that this virtual system indeed has $q=x$ and $q=x_d$ as its particular solutions).
\end{theorem}
\begin{proof}
Let $V = \delta q^{\top}M\delta q$. Using \eqref{Vsl_LV} computed in the proof of Theorem~\ref{Thm:learning1}, along with the bound on $d_q$ \eqref{d_bound} and the contraction condition \eqref{eq_MdotContracting} for $\Xi = 2\alpha M$, we have that
\begin{align}
\label{Vdot_general}
\dot{V} \leq -2\alpha V+2\sqrt{\overline{m}}(L_u(\epsilon_{\ell0}+\epsilon_{\ell1}V_{\ell}/\sqrt{\underline{m}})+\bar{d})\|\Theta \delta q\|
\end{align}
where the relation $\int^x_{x_d}\|\delta q\|\leq V_{\ell}/\sqrt{\underline{m}}$ is used. Since we have $\dot{V} = \|\Theta \delta q\|(d\|\Theta \delta q\|/dt)$ and $V=\|\Theta \delta q\|^2$, \eqref{Vdot_general} implies that $\dot{V}_{\ell} \leq -(\alpha-L_u\epsilon_{\ell1}\sqrt{\overline{m}/\underline{m}})V_{\ell}+\sqrt{\overline{m}}(L_u\epsilon_{\ell0}+\bar{d})$, which results in \eqref{bound_no_dist} by applying the comparison lemma~\cite[pp.102-103, pp.350-353]{Khalil:1173048} as long as the condition \eqref{epsilon_ell2_condition_det} holds.
\end{proof}
\section{A Convex Optimization Approach for NCM Robust Control~\cite{ncm,nscm,ancm,cvstem,tutorial}}
\label{Sec:ncm}
Although Theorems~\ref{Thm:learning1} and~\ref{Cor:learning_simple} are already useful for systems with a known contraction metric and corresponding differential Lyapunov function, finding them for general nonlinear systems is not always straightforward. This section delineates one approach, called ConVex optimization-based Steady-state Tracking Error Minimization (CV-STEM), to optimally design $u^*$ and $M$ of Theorem~\ref{Thm:learning1} for control-affine nonlinear systems, thereby demonstrating how to construct the NCM of Definition~\ref{Def:NCM} in practice for probably stable learning-based robust control.

Let us consider the following control-affine nonlinear system:
\begin{align}
\label{affine_dynamics}
d{x}=&({f}({x},t)+B(x,t)u)dt+d(x,t)dt+G(x,t)d\mathscr{W}(t)
\end{align}
where $t$, $x$, $u$, $d$, $G$, and $\mathscr{W}$ are as defined in \eqref{eq:xfx}, and ${f}:
\mathbb{R}^n\times\mathbb{R}_{+}\rightarrow\mathbb{R}^n$ and $B:\mathbb{R}^n\times\mathbb{R}_{+} \to \mathbb{R}^{n\times m}$ are known smooth functions.
\begin{remark}
If ${h}(x,u,t)-({f}({x},t)+B(x,t)u)$ is bounded in \eqref{eq:xfx} and \eqref{affine_dynamics}, we can still apply the technique to be discussed in this section even for control non-affine systems \eqref{eq:xfx}, by treating this modeling error as external disturbance $d$ in \eqref{affine_dynamics}. For situations where such an assumption is not valid, we could apply (learning-based) adaptive control~\cite{9109296,lopez2021universal,ancm,8794351} to estimate the non-affine part of the dynamics.
\end{remark}
\subsection{Contraction Theory-based Robust Control (CV-STEM)}
\label{sec_cvstem_robust}
We design a contracting control law $u=u^*$ for \eqref{affine_dynamics} by the following differential feedback control law~\cite{ccm} (see Remark~\ref{control_form_remark} for more options):
\begin{align}
\label{differential_u}
&u^*(x,x_d,u_d,t) = u_d-\int_{x_d}^xR(q,t)^{-1}B(q,t)^{\top}M(q,t)\delta q
\end{align}
where $R(x,t) \succ 0$ is a given weight matrix, $M(x,t) \succ 0$ is a contraction metric to be defined in Theorem~\ref{Thm:cvstem_ccm}, $q$ is a smooth path that connects $x$ to $x_d$ as in \eqref{virtual_system_general}, and $x_d$ is a given target trajectory of \eqref{target_dynamics} with ${h}(x,u,t)=f(x,t)+B(x,t)u$. Note that \eqref{differential_u} yields $\delta u^*=-R(x,t)^{-1}B(x,t)^{\top}M(x,t)\delta x$.
\begin{remark}
\label{control_form_remark}
We could utilize other types of control laws in \eqref{differential_u}, \eg{}, $\delta u = k(x,\delta x, u, t)$ (see Sec.~\ref{Sec:cosynthesis})~\cite{ccm} or $u=u_d-K(x,x_d,u_d,t)(x-x_d)$~\cite{cvstem,ncm,nscm,ancm}. See~\cite{tutorial} for their trade-offs.
\end{remark}

The CV-STEM is for designing $u^*$ and $M$ of Theorem~\ref{Thm:learning1} explicitly by \eqref{differential_u} to minimize the steady-state bound of \eqref{Eq:learning_bound_general}.
\begin{theorem}
\label{Thm:cvstem_ccm}
Suppose that the contraction metric $M(x,t) = W(x,t)^{-1} \succ 0$ of \eqref{differential_u} is designed by the following convex optimization (CV-STEM)~\cite{ncm,cvstem,nscm,tutorial}, which minimizes the steady-state upper bound of \eqref{Eq:learning_bound_general}:
\begin{align}
    \label{convex_opt_ccm}
    &{J}_{CV}^* = \min_{\nu >0,\chi \in \mathbb{R},\bar{W}\succ 0} \frac{C}{2\alpha_{\ell}}\chi \text{~~\st{} \eqref{deterministic_contraction_tilde}, \eqref{deterministic_contraction_tilde2}, and \eqref{W_tilde}}
\end{align}
with the convex constraints \eqref{deterministic_contraction_tilde}, \eqref{deterministic_contraction_tilde2}, and \eqref{W_tilde} given as
\begin{align}
\label{deterministic_contraction_tilde}
&\begin{bmatrix}H(\nu,\bar{W})+2\alpha \bar{W} && \bar{W} \\ \bar{W} && \nu\alpha_s^{-1}I \end{bmatrix}  \preceq 0,~\forall x,t \\
\label{deterministic_contraction_tilde2}
&-\partial_{b_i}\bar{W}+2\sym{}\left(\frac{\partial b_i}{\partial x}\bar{W}\right) = 0,~\forall i=1,\cdots,m,~\forall x,t \\
\label{W_tilde}
&I \preceq \bar{W}(x,t) \preceq \chi I,~\forall x,t
\end{align}
where $\partial_{p} F = \sum_{k=1}^n(\partial F/\partial x_k)p_k$ for $p(x,t)\in\mathbb{R}^n$ and $F(x,t) \in \mathbb{R}^{n\times n}$, $\nu = \overline{m}$, $\chi = \overline{m}/\underline{m}$, $\bar{W} = \nu W$, $f$ and $B$ are given in \eqref{affine_dynamics}, $b_i$ is the $i$th column of $B$, and
\begin{align}
H(\nu,\bar{W}) = -\frac{\partial \bar{W}}{\partial t}-\partial_f \bar{W}+2\sym{}\left(\frac{\partial f}{\partial x}\bar{W}\right)-\nu BR^{-1}B^{\top}.
\end{align}
with the other variables defined in Theorem~\ref{Thm:learning1}. If $u^*$ is designed as \eqref{differential_u} for $M$ of \eqref{convex_opt_ccm}, the contraction condition \eqref{eq_MdotContracting} of Definition~\ref{def:contracting} and \eqref{Mcon} of Theorem~\ref{Thm:learning1} hold with $\Xi$ given by \eqref{def:Xi}, ${h}(x,u,t)=f(x,t)+B(x,t)u$, and $u=u^*$.
\end{theorem}
\begin{proof}
Applying Schur's complement lemma~\cite[pp. 7]{lmi} to the constraint \eqref{deterministic_contraction_tilde}, we have that
\begin{align}
\label{schur_tilde}
H(\nu,\bar{W})+2\alpha\bar{W}-\alpha_s\bar{W}^2/\nu \preceq 0.
\end{align}
It can be easily verified that, by multiplying the constraints \eqref{schur_tilde}, \eqref{deterministic_contraction_tilde2}, and \eqref{W_tilde} by $M=W^{-1}$ from both sides and by $\nu^{-1}$, they can be equivalently rewritten as~\cite{cvstem}
\begin{align}
\label{ccm1}
&\frac{\partial M}{\partial t}+\partial_f M+2\sym{}(M({\partial f}/{\partial x}))-MBR^{-1}B^{\top}M \preceq -\Xi \\
\label{ccm2}
&\partial_{b_i}M+2\sym{}(M({\partial b_i}/{\partial x})) = 0 \\
\label{ccm3}
&\underline{m}I \preceq M \preceq \overline{m}I
\end{align}
respectively, for $\Xi$ of \eqref{def:Xi}, where \eqref{ccm3} indicates that \eqref{W_tilde} is indeed equivalent to \eqref{Mcon}. Combining \eqref{ccm1} and \eqref{ccm2}, we get
\begin{align}
\dot{M}+2\sym{}(M({\partial (f(x,t)+B(x,t)u^*)}/{\partial x})) \preceq -\Xi
\end{align}
which implies the contraction condition \eqref{eq_MdotContracting} holds for ${h}(x,u,t)=f(x,t)+B(x,t)u$ and $u=u^*$.

Furthermore, it can be seen that the problem \eqref{convex_opt_ccm} minimizes the steady-state upper bound of \eqref{Eq:learning_bound_general} due to the relation $\lim_{t\to\infty}\mathop{\mathbb{E}}\left[\|x(t)-x_d(t)\|^2\right] = (C/(2\alpha_{\ell}))(\overline{m}/\underline{m}) = (C/(2\alpha_{\ell}))\chi$. Note that \eqref{convex_opt_ccm} is convex as the objective is affine in $\chi$, and \eqref{deterministic_contraction_tilde}, \eqref{deterministic_contraction_tilde2}, and \eqref{W_tilde} are linear matrix inequalities in terms of $\nu$, $\chi$, and $\bar{W}$~\cite[pp. 7]{lmi}.
\end{proof}
\subsection{NCM Construction using CV-STEM}
Now that we can construct the contracting control law $u^*$ of Theorem~\ref{Thm:learning1} by using Theorem~\ref{Thm:cvstem_ccm}, we are ready to explicitly design the NCM of Definition~\ref{Def:NCM}. Since the NCM only requires one DNN evaluation at each time instant without solving any optimization problems unlike the CV-STEM of Theorem~\ref{Thm:cvstem_ccm}, it enables real-time computation of optimal feedback control using \eqref{differential_u} in most engineering and science applications. We formally derive its robustness and stability guarantees in the following theorems.
\begin{theorem}
\label{Thm:NCMstability1}
Let $u_L$ be a learning-based controller, prameterized by the NCM of Definition~\ref{Def:NCM}, which approximates the contracting control law $u^*$ of \eqref{differential_u}. If Assumption~\ref{assump_learning} holds for such $u_L$, $u^*$, and $M$ of Theorem~\ref{Thm:cvstem_ccm} with ${h}(x,u,t) =f(x,t)+B(x,t)u$ of \eqref{affine_dynamics}, then Theorem~\ref{Thm:learning1} holds, \ie{}, the system trajectory of \eqref{affine_dynamics} controlled by $u_L$ exponentially converges to a ball around the target trajectory $x_d$ of \eqref{target_dynamics} with ${h}(x,u,t) =f(x,t)+B(x,t)u$.
\end{theorem}
\begin{proof}
Since $u^*$ of \eqref{differential_u} with $M$ of \eqref{convex_opt_ccm} guarantees the assumptions of Theorem~\ref{Thm:learning1} due to Theorem~\ref{Thm:cvstem_ccm}, the exponential bound \eqref{Eq:learning_bound_general} holds as long as Assumption~\ref{assump_learning} is satisfied.
\end{proof}

Although Theorem~\ref{Thm:NCMstability1} considers the modeling error on the learned control law, $\|u_L-u^*\|$, it is also useful to study how the learning error on the contraction metric, $\|M_L-M\|$, affects the NCM stability performance.
\begin{figure}
    \centering
    \includegraphics[width=85mm]{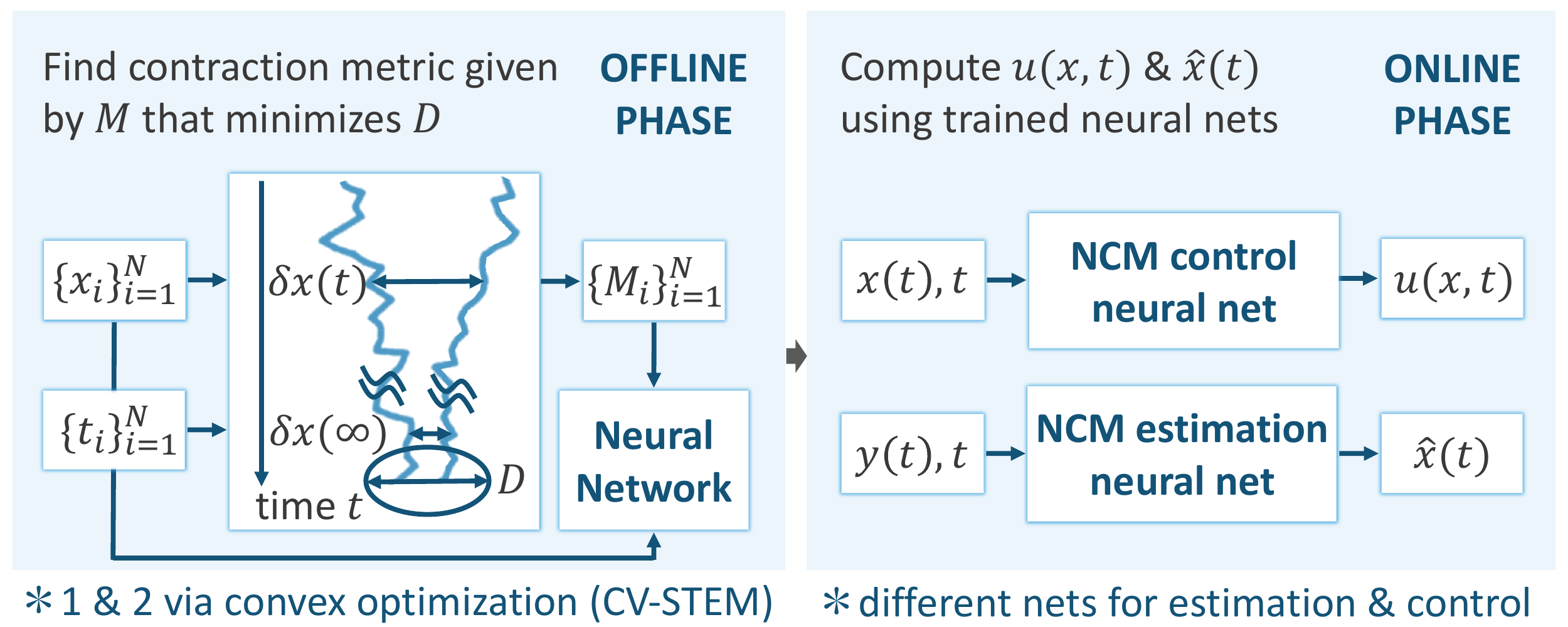}
    \vspace{-0.8em}
    \caption{Illustration of NCM ($x$: system state; $M$: positive definite matrix that defines optimal contraction metric; $x_i$ and $M_i$: sampled $x$ and $M$; $\hat{x}$: estimated system state; $y$: measurement; and $u$: system control input). Note that the target trajectory $(x_d,u_d)$ is omitted in the figure fore simplicity. (see~\cite{ncm,tutorial} for state estimation).}
    \label{ncmdrawing}
\vspace{-1.5em}
\end{figure}
\begin{theorem}
\label{Thm:NCMstability2}
Let $M_L$ be the NCM of Definition~\ref{Def:NCM} that models $M$ of Theorem~\ref{Thm:cvstem_ccm}. Suppose that \eqref{eq_Mlipschitz} holds for such $M$, and that the system \eqref{affine_dynamics} is controlled by \eqref{differential_u} with $M$ replaced by $M_L$. Suppose also that $\exists \bar{b},\bar{\rho}\in[0,\infty)$ \st{} $\|B(x,t)\|\leq\bar{b}$ and $\|R^{-1}(x,t)\|\leq\bar{\rho},~\forall (x,t)\in{S}_x\times{S}_t$, for $B$ in \eqref{affine_dynamics} and $R$ in \eqref{differential_u}, where ${S}_x$ and ${S}_t$ are given compact sets of \eqref{learning_cond}. If the NCM satisfies
\begin{align}
\label{Merror}
\|M_L(x,t)-M(x,t)\| \leq \epsilon_{\ell},~\forall (x,t)\in{S}_x\times{S}_t
\end{align}
for $\exists \epsilon_{\ell}\in[0,\infty)$, Theorem~\ref{Thm:learning1} holds for $\epsilon_{\ell0}=0$ and $\epsilon_{\ell1}=\bar{\rho}\bar{b}^2\epsilon_{\ell}$.
\end{theorem}
\begin{proof}
This follows from Theorem~\ref{Thm:NCMstability1}, due to the facts that the learning error of \eqref{learning_cond} is given as $\|u_L-u^*\|\leq\bar{\rho}\bar{b}\epsilon_{\ell}\int^{x}_{x_d}\|\delta q\|$ due to \eqref{differential_u}, and the Lipschitz constant of \eqref{Lipschitz_u} is as $L_u=\bar{b}$.
\end{proof}

Theorems~\ref{Thm:NCMstability1} and~\ref{Thm:NCMstability2} indeed guarantee robustness and incremental stability of the NCM learning-based control law $u_L$, which imitates the contraction theory-based robust control $u^*$ of Theorem~\ref{Thm:cvstem_ccm} for its real-time implementability.
\section{Learning Certified Control using Contraction Metrics~\cite{chuchu}}
\label{Sec:cosynthesis}
This section delineates one framework to train a DNN for jointly synthesizing a contraction metric and contraction control law with the robustness and stability guarantees of Theorems~\ref{Thm:NCMstability1} and~\ref{Thm:NCMstability2}, treating the contraction constraints \eqref{eq_MdotContracting} and \eqref{Mcon} as the loss functions. Intuitively, this can be viewed as a way to solve \eqref{convex_opt_ccm} of Theorem~\ref{Thm:cvstem_ccm} directly by a DNN, while also avoiding the integral evaluation required in \eqref{differential_u}.
\begin{remark}
Although here we consider the system \eqref{affine_dynamics} with $G=0$, stochastic perturbation could be incorporated in a similar way using the constraints introduced in \eqref{deterministic_contraction_tilde} and \eqref{deterministic_contraction_tilde2}.
\end{remark}

For deterministic systems \eqref{affine_dynamics} with $G=0$, it is shown in~\cite{ccm} that the following conditions weaker than \eqref{deterministic_contraction_tilde} and \eqref{deterministic_contraction_tilde2} are sufficient for the existence of a contracting differential feedback controller, which satisfies the contraction condition \eqref{eq_MdotContracting} of Definition~\ref{def:contracting} with $\Xi = 2\alpha M = 2\alpha W^{-1}$:
\begin{align}
&B_{\bot}^\T \left( -\frac{\partial W}{\partial t}-\partial_f W + 2\sym{}\left({\frac{\partial f}{\partial x} W}\right) + 2 \alpha W \right)B_{\bot} \prec 0\label{eq:c1} \\
&B_{\bot}^\T \left( \partial_{b_i} W - 2\sym{}\left({\frac{\partial b_i}{\partial x} W}\right) \right) B_{\bot} = 0,\forall j,x,t
\label{eq:c2}
\end{align}
where $B_{\bot}(x,t)$ is an annihilator matrix of $B(x,t)$ satisfying $B_{\bot}B$= 0, and the other notations and variables are as defined in Theorem~\ref{Thm:cvstem_ccm}. Thus, in this section, we consider the constraints \eqref{eq:c1} and \eqref{eq:c2}, instead of \eqref{deterministic_contraction_tilde} and \eqref{deterministic_contraction_tilde2}, without specifying the form of the contracting control $u^*$ as in \eqref{differential_u}.
\subsection{Cost Functions in Neural Network Training}
The purpose of this section is to find a DNN-based controller $u_L(x,x_d,u_d,t; \theta_u)$ that models a contracting control law $u^*$ and dual contraction metric $W_L$ designed as
\begin{align}
\label{Def:dualWnet}
W_L(x,t;\theta_w)=\Theta_L(x,t;\theta_{\vartheta})^{\top}\Theta_L(x,t;\theta_{\vartheta})+\overline{m}_L^{-1}I
\end{align}
that models $W=M^{-1}$ of Theorem~\ref{Thm:cvstem_ccm}, where $\theta_u$ and $\theta_w = \{\theta_{\vartheta},\overline{m}_L\}$ with $\overline{m}_L>0$ are hyper-parameters. Note that by definition of $W_L$ in \eqref{Def:dualWnet}, it is already a positive definite matrix satisfying $W_L(x,t;\theta_w) \succeq \overline{m}_L^{-1}I$ as in \eqref{Mcon} or \eqref{W_tilde}. 

Let $C_u(x, x_d, u_d, t; \theta_w, \theta_u)$ be the left-hand side of \eqref{eq_MdotContracting} with $W(= M^{-1}) = W_L$, $u^*=u_L$, ${h}(x,u,t)=f(x,t)+B(x,t)u$, and $\Xi=2\alpha M$, and $\rho({S})$ be the uniform distribution over the compact set of \eqref{learning_cond}. We define the contraction loss as follows:
\begin{equation}
{\mathcal{L}}_u(\theta_w, \theta_u) = \mathop{\mathbb{E}}_{(x,x_d,u_d,t) \sim \rho({S})}{L_{PD}(-C_u(x, x_d, u_d, t;\theta_w, \theta_u))}
\label{eq:loss_controller}
\end{equation}
where $L_{PD}(\cdot) \geq 0$ is a function for penalizing non-positive definiteness, \ie{}, $L_{PD}(A)=0$ if and only if $A \succeq 0$. Also, since we already have $W_L \succeq \overline{m}_L^{-1}I$ by \eqref{Def:dualWnet}, the condition \eqref{Mcon} on the boundedness of $W_L$ can be considered by the following loss function with another hyper-parameter $\underline{m}_L$ in $\theta_w$:
\begin{equation}
{\mathcal{L}}_c(\theta_w) = \mathop{\mathbb{E}}_{(x,x_d,u_d,t) \sim \rho({S})}{L_{PD}(\underline{m}_L^{-1}I - W_L(x,t;\theta_w))}.
\label{eq:loss_condition_number}
\end{equation}
Furthermore, we utilize the weaker contraction conditions \eqref{eq:c1} and \eqref{eq:c2} also as the following loss functions to provide more guidance for DNN optimization, as simultaneously learning $u_L$ and $W_L$ by minimizing ${\mathcal{L}}_u$ and $\mathcal{L}_c$ solely is challenging as demonstrated in~\cite{chuchu}:
\begin{align}\label{eq:loss_metric}
{\mathcal{L}}_{w1}(\theta_w) =& \mathop{\mathbb{E}}\limits_{(x,x_d,u_d,t) \sim \rho({S})} L_{PD}(-C_1(x,t;\theta_w)) \\
{\mathcal{L}}_{w2}(\theta_w) =& \sum_{j=1}^{m} \mathop{\mathbb{E}}\limits_{(x,x_d,u_d,t) \sim \rho({S})} \|C_2^i(x,t;\theta_w)\|_F\label{eq:B_regu}
\end{align}
where $C_1(x,t; \theta_w)$ and $\{C_2^j(x,t; \theta_w)\}_{j=1}^{m}$ are the left-hand side of \eqref{eq:c1} and \eqref{eq:c2} with $W=W_L$, respectively.

Approximating the expectations of \eqref{eq:loss_controller}--\eqref{eq:B_regu} using sampled data, the empirical loss function for the NCM training is defined as follows:
\begin{align}
\label{loss_func}
&{\mathcal{L}}(\theta_u,\theta_w) = \frac{1}{N} \sum_{i=1}^{N} \mathcal{L}_{m,i}(\theta_u,\theta_w)+\mathcal{L}_{w,i}(\theta_w) \\
&{\mathcal{L}}_{m,i} = L_{PD}(-C_u(\xi_i;\theta_w, \theta_u))+L_{PD}(\underline{m}_L^{-1}I - W(x_i,t_i;\theta_w)) \nonumber \\
&{\mathcal{L}}_{w,i} = L_{PD}(-C_1(x_i,t_i;\theta_w)) + \sum_{j=1}^{m} \|C_2^j(x_i,t_i;\theta_w))\|_F
\end{align}
where the training samples $\{\xi_i=(x_i, x_{d,i}, u_{d,i}, t_i)\}_{i=1}^{N}$ are drawn independently from $\rho({S})$, and $L_{PD}$ is computed as $L_{PD}(A) = \frac{1}{K} \sum_{i=1}^{K} \min\{0, -p_i ^\T A p_i\}$ for a given matrix $A \in \reals^{n \times n}$ and randomly sampled $K$ points in the set $\{p_i \in \reals^n~|~\|p_i\| = 1\}_{i=1}^{K}$. The robustness and stability guarantee of $u_L$ is given in the following theorem.
\begin{theorem}
\label{Thm:cosyn}
Suppose Assumption~\ref{assump_learning} holds for the DNN-based NCM controller $u_L$ learned by minimizing the loss function \eqref{loss_func}. If \eqref{affine_dynamics} with $G=0$ is controlled by $u_L$, \eqref{bound_no_dist} of Theorem~\ref{Cor:learning_simple} holds, \ie{}, the distance between the target trajectory $x_d$ and perturbed trajectories $x$ is exponentially bounded.
\end{theorem}
\begin{proof}
This follows from the proof of Theorem~\ref{Cor:learning_simple} as long as $u_L$ is learned to satisfy \eqref{learning_cond} with the loss function \eqref{loss_func}.
\end{proof}
\begin{remark}
The objective function \eqref{convex_opt_ccm} of the CV-STEM method in Theorem~\ref{Thm:cvstem_ccm} can also be considered in \eqref{loss_func} as a loss function ${\mathcal{L}}_{CV}(\theta_w) = ({C}/{2\alpha_{\ell}})({\overline{m}_L}/{\underline{m}_L})$, where $\overline{m}_L$ and $\underline{m}_L$ are given in \eqref{Def:dualWnet} and \eqref{eq:loss_condition_number}. Using it in \eqref{loss_func} would augment $u_L$ of Theorem~\ref{Thm:cosyn} with some CV-STEM-type optimality property.
\end{remark}
\subsection{Numerical Simulation}
We briefly summarize the simulation results presented in~\cite{chuchu} to see how the theoretical guarantees in Theorems~\ref{Thm:learning1}--\ref{Thm:cosyn} lead to exponential stabilization of nonlinear systems in practice (more simulation results can be found also in~\cite{ncm,nscm,ancm,lagros,tutorial}). The loss function~\eqref{loss_func} is used for DNN training, and we consider \begin{inparaenum}[(a)]
    \item {PVTOL}: Planar Vertical-Takeoff-Vertical-Landing (PVTOL) system for drones~\cite{7989693, 47710};
    \item {Quadrotor}: Physical quadrotor system~\cite{7989693};
    \item {Neural lander}: Drone flying close to the ground, where the ground effect is non-negligible~\cite{8794351} (the ground effect is learned from empirical data using a $4$-layer neural network); and
    \item {SEGWAY}: Real-world Segway robot~\cite{taylor2019learning}. 
\end{inparaenum}

The performance is compared with the Sum of Squares (SoS)-based method~\cite{7989693}, Model Predictive Control (MPC)~\cite{mpcPytorch,tassa2014control}, and proximal policy optimization reinforcement learning (RL) algorithm~\cite{schulman2017proximal}.
\subsubsection{DNN Training}
For $\Theta_L$ of the dual metric in \eqref{Def:dualWnet}, we use a $2$-layer neural network with $128$ neurons. For $u_L$ of Theorem~\ref{Thm:cosyn}, we use  $u_L(x, x_d, u_d, t; \theta_u) = w_2  \cdot \tanh(w_1 \cdot (x-x_d)) + u_d$ where $w_1 = w_1(x, x_d; \theta_{u1})$ and $w_2 = w_2(x, x_d; \theta_{u2})$ are two 2-layer neural networks with 128 neurons with the hyperparameters $\theta_u = \{\theta_{u1},\theta_{u2}\}$. We train the networks for $20$ epochs with a training set with 130K samples. See~\cite{chuchu,ncm,nscm,lagros} for more details on the NCM dataset generation.
\subsubsection{Target Trajectory Generation}
The target control inputs $u_d(t)$ are generated as the linear combination of the elements in a set of sinusoidal signals with some fixed frequencies and randomly sampled a weight for each frequency component. The initial states of the target trajectories $x_d(0)$ are uniformly randomly sampled from a compact set, and then $x_d$ are obtained by integrating~\eqref{target_dynamics}. The initial errors $\mathtt{e}(0)$ are uniformly randomly sampled from a compact set, and the initial states $x(0)$ are computed as $x(0) = x_d(0) + \mathtt{e}(0)$.
\begin{figure}
\centering
    \includegraphics[width=0.2\textwidth,trim=0 0 0 0,clip]{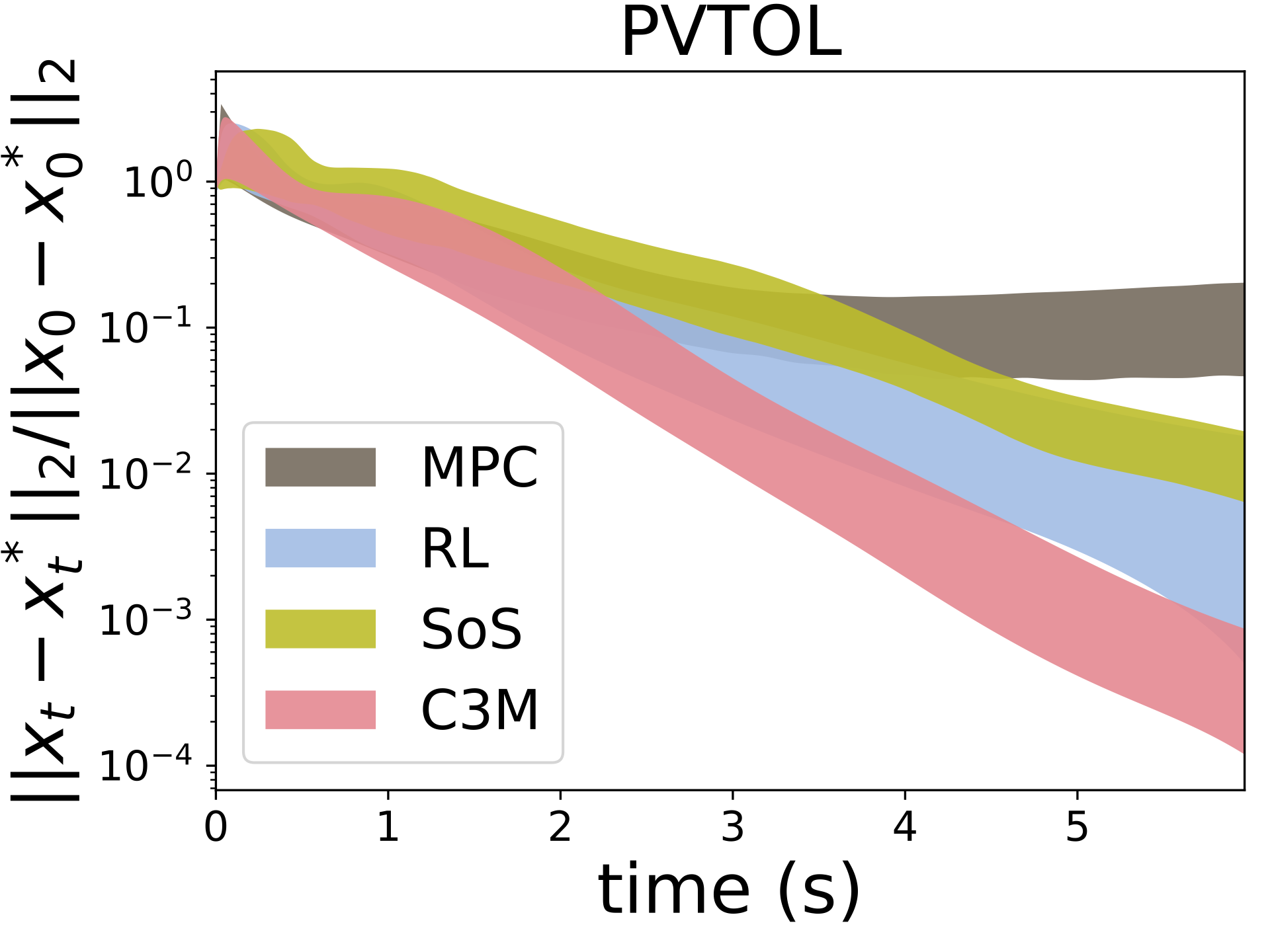}
    \includegraphics[width=0.2\textwidth,trim=0 0 0 0,clip]{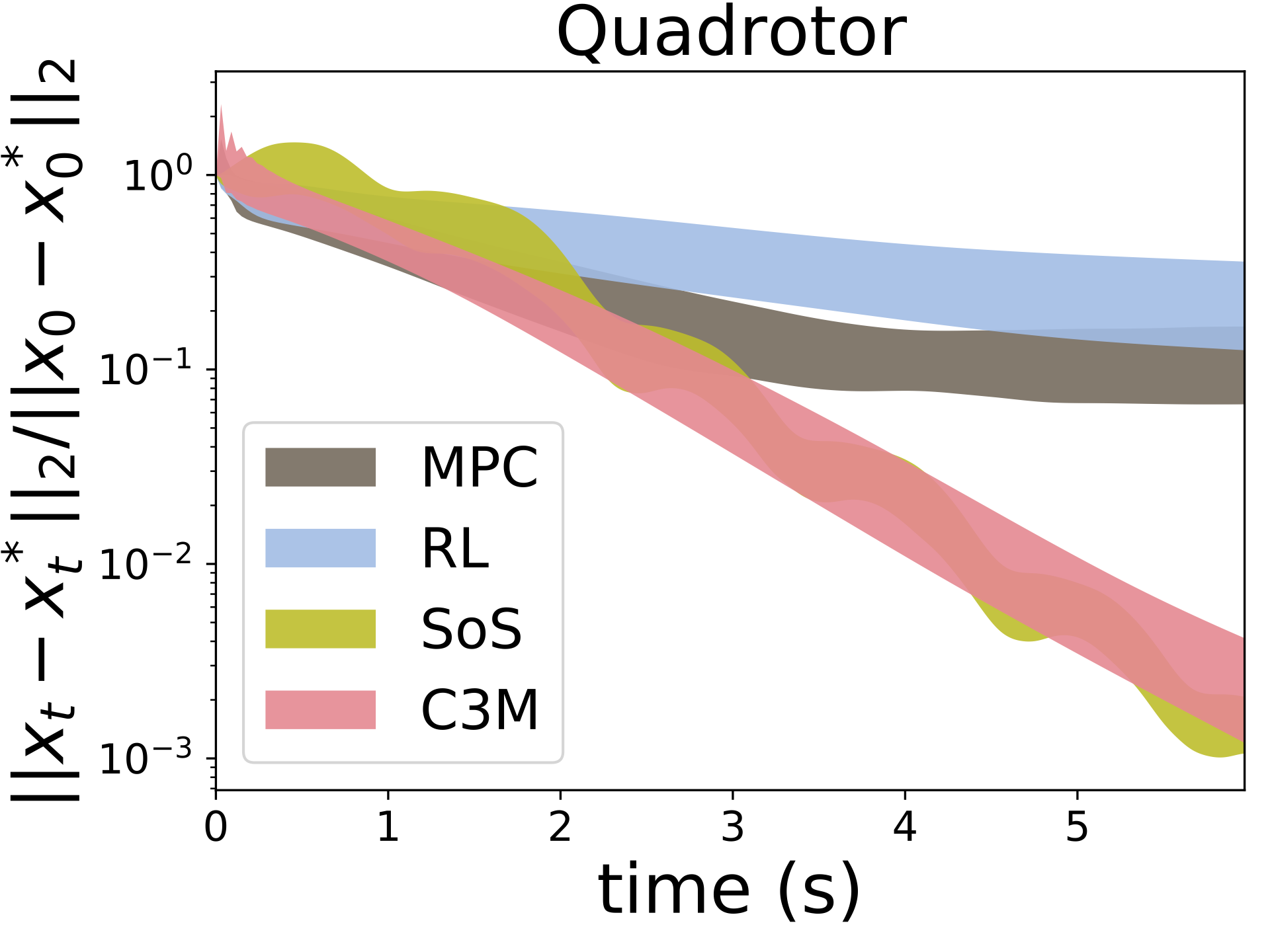}
    \includegraphics[width=0.2\textwidth,trim=0 0 0 0,clip]{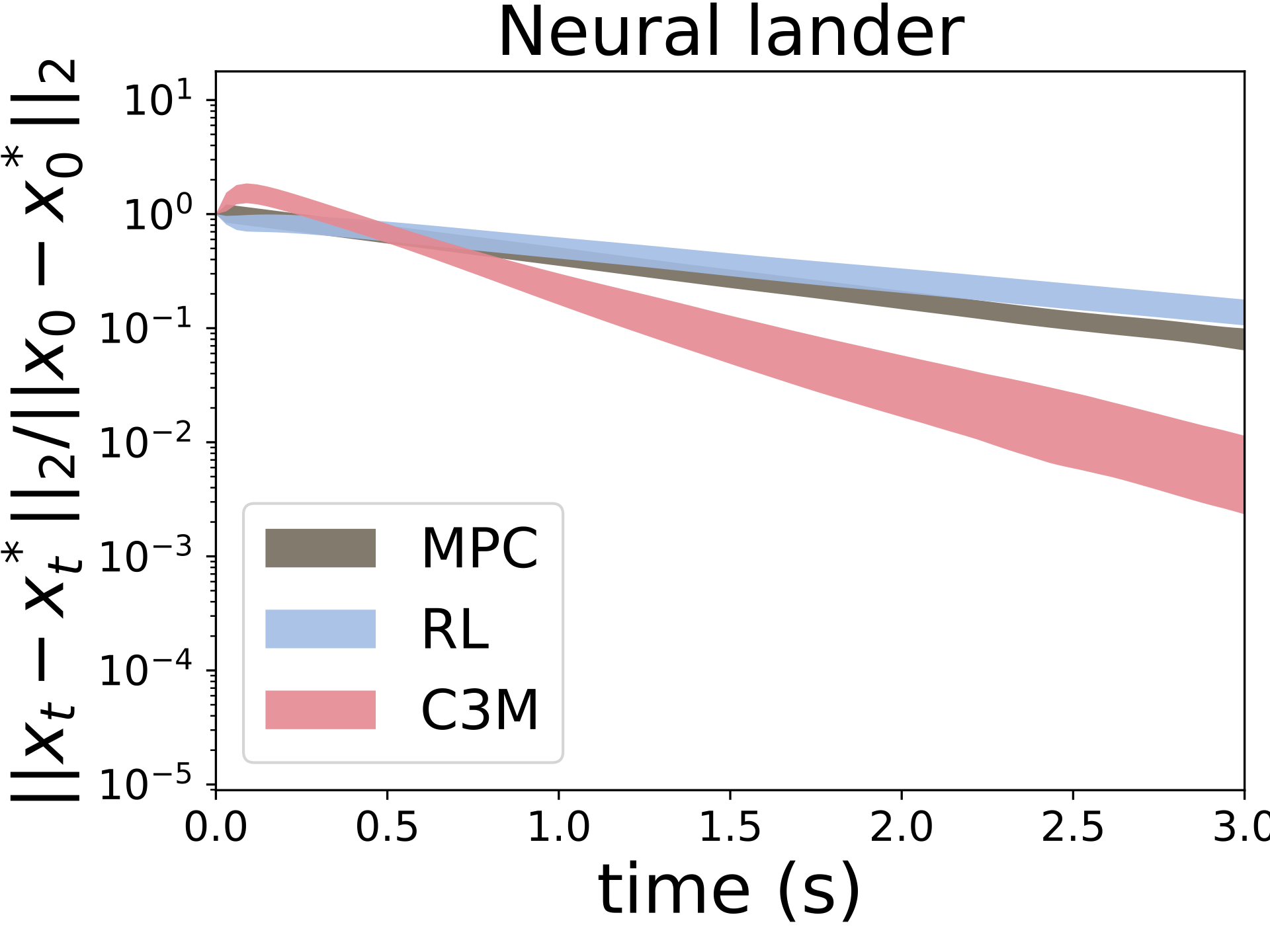}    \includegraphics[width=0.2\textwidth,trim=0 0 0 0,clip]{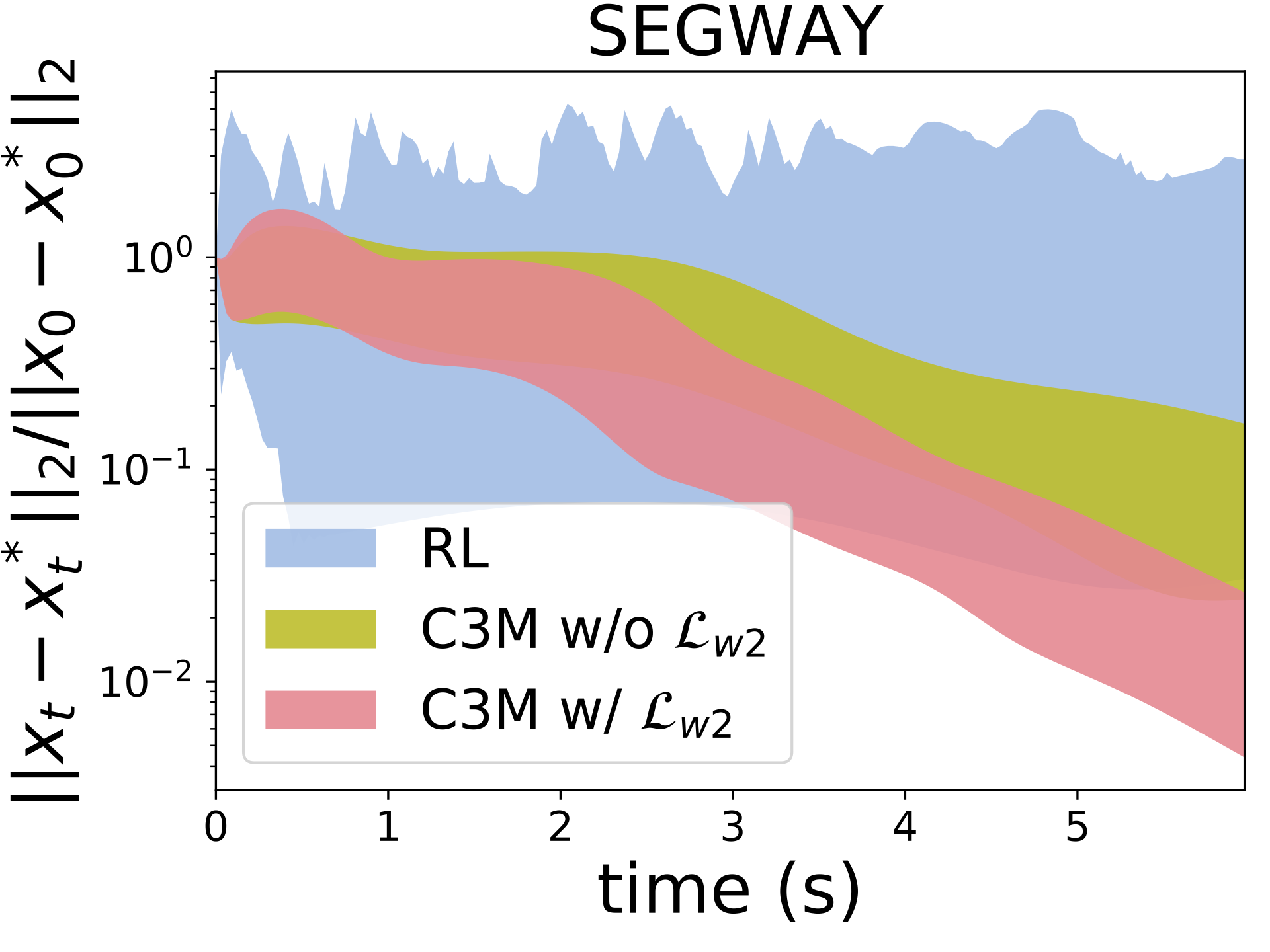}
\vspace{-0.8em}
\caption{\footnotesize Normalized tracking error on the benchmarks using different approaches. The $y$ axes are in log-scale. The tubes are tracking errors between mean plus and minus one standard deviation over $100$ trajectories (C3M: Proposed approach in Theorem~\ref{Thm:cosyn}).}
\label{fig:results}
\vspace{-1.5em}
\end{figure}
\subsubsection{Simulation Results and Discussions~\cite{chuchu}}
Figure~\ref{fig:results} shows the normalized tracking error defined as $x_e(t) = ||x(t) - x_d(t)||/ ||x(0) - x_d(0)||$ for each model and method. For PVTOL and the quadrotor models, both SoS and our learning-based method find a contracting tracking controller that renders $x_e$ decrease rapidly, successfully achieving exponential stabilization. While the SoS of~\cite{7989693} cannot solve for the Neural lander and SEGWAY examples due to their complex and non-polynomial dynamics, and MPC produces large numerical errors for the SEGWAY example due to the sensitivity of the SEGWAY model, the proposed control of Theorem~\ref{Thm:cosyn} yields consistently higher tracking performance for both of these models. The performance of the RL method varies with tasks: it achieves comparable results for PVTOL and the neural lander models but fails to find a contracting tracking controller within a reasonable time for the quadrotor and Segway models, as their state spaces are larger than the previous two cases.

Figure~\ref{fig:results} thus implies that our proposed method outperforms both the learning and non-learning-based methods, producing a contracting controller with smaller tracking errors. These errors are guaranteed to be bounded exponentially with time even with the learning error and external disturbances, as rigorously proven in Theorems~\ref{Thm:learning1}--\ref{Thm:cosyn}. Again, more simulation results can be found in~\cite{ncm,nscm,ancm,lagros,tutorial}.
\section{Conclusion}
\label{Sec:conclusion}
In this paper, we presented a theoretical overview of the NCM for provably stable, robust, and optimal learning-based control of non-autonomous nonlinear systems. It is formally shown that, even under the presence of learning errors and deterministic/stochastic disturbances, the NCM control ensures the distance between a target trajectory and learned system trajectories to be bounded exponentially with time, enabling safe and robust use of AL and ML-based automatic control frameworks in real-world scenarios. The numerical simulation results of~\cite{chuchu} are presented to validate the effectiveness of the NCM framework.

\bibliographystyle{IEEEtran}
\bibliography{root}

\end{document}